\documentclass[10pt,twocolumn,letterpaper]{article}

\usepackage[preprint]{cvpr}
\usepackage{pratik}

\usepackage[color=gray!30,textsize=tiny]{todonotes}

\newlength{\mylen}
\makeatletter
\renewcommand\paragraph{\@startsection{paragraph}{4}{\z@}{1.25ex \@plus 0.5ex \@minus 0.2ex}{-1em}{\normalfont\normalsize\bfseries}}
\DeclareMathSizes{9.95}{9}{7}{5}
\DeclareMathSizes{10}{9}{7}{5}
\DeclareMathSizes{11}{10}{8}{6}
\makeatother

\def\confName{CVPR}
\def\confYear{2027}

\title{
Visual Thinking: Training-Free Active Perception for\\ Counting and Localizing Hard-to-Find Objects
}

\author{Emil Kielar and Pratik Chaudhari\\[0.25em]
University of Pennsylvania\\[0.25em]
Email: \href{ekielar@upenn.edu}{ekielar@upenn.edu}, \href{pratikac@upenn.edu}{pratikac@upenn.edu}
}

\begin{document}

\maketitle

\begin{abstract}
Vision foundation models can precisely segment objects based solely on a text prompt. However, they struggle to recover objects that are small, occluded or whose appearance is poorly captured by the original textual description. Vision language models can examine the image and its context, they can propose richer descriptions of the objects to be detected. But these models struggle with precise segmentation and localization. We introduce Visual Thinking, a training-free framework that combines a vision-language model with Segment Anything Model 3 (SAM 3) through an iterative active perception loop. At each iteration, the VLM proposes short noun phrases using the image, candidate crops and history from past iterations. A controller ranks these phrases using embedding similarity between the original prompt and the proposal, an entropy-based verification priority and the novelty of the prompt. Next, SAM 3 uses prompts to detect objects in the image. We associate detections across iterations and update each object's confidence score using a weighted mean of scores obtained from SAM 3 with a larger weight for detections using narrower prompts. We supply confident detections as positive visual exemplars for SAM~3, and incorporate image tiling to recover small objects in cluttered images. We evaluate counting and localization on CARPK, FSCD-147, PixMo-Count, and OmniCount (Fruits). Compared with SAM 3, Visual Thinking reduces the counting mean absolute error (MAE) by about 18 on FSC-147 and about 8 on CARPK. It improves box average precision (AP) by more than 12 on FSCD-147 and CARPK. Visual Thinking, with fixed or no tiling, performs on par with Claude Opus 5.5 at extra-high reasoning effort, while running on single desktop GPU in about 5.91 seconds per image, compared to about 46.53 seconds per image for Claude through its API.
\end{abstract}

\section{Introduction}
\label{s:intro}

\begin{figure}[t]
    \centering
    \includegraphics[width=\linewidth]{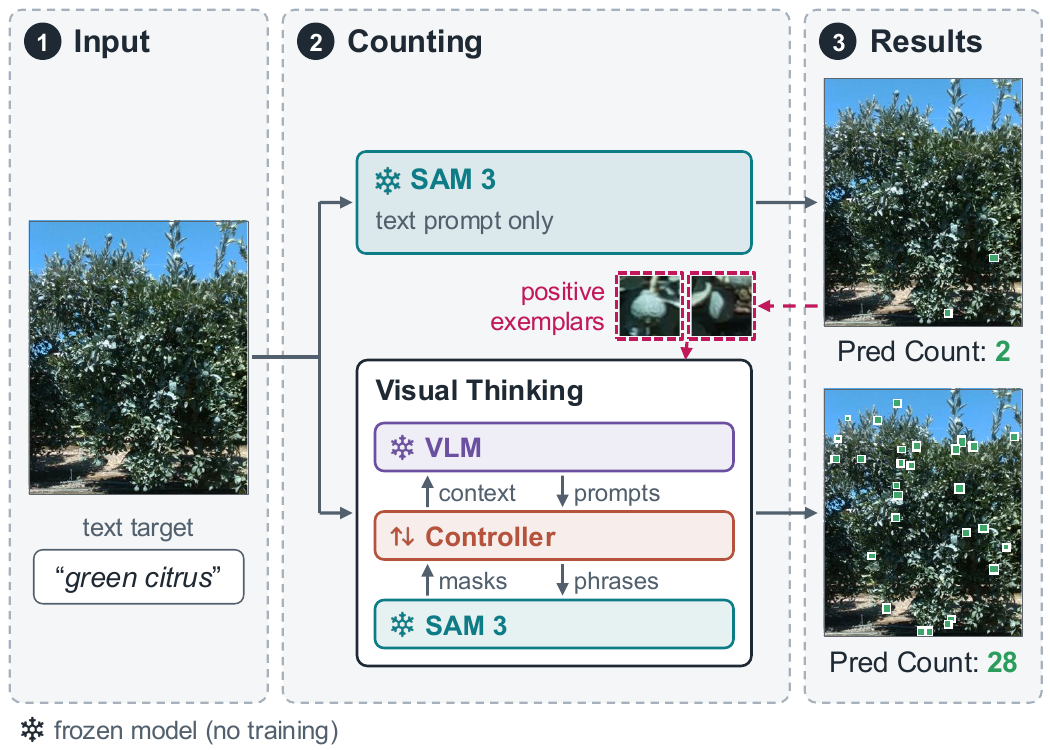}
    \caption{\textbf{Visual Thinking} turns SAM~3 and a VLM into an iterative active perception loop. The VLM proposes new text prompts, and the controller ranks them. Next, the prompts are run on SAM~3, utilizing confident detections as positive exemplars. Masks from all searches are saved and deduplicated in an object library, from which the final count is calculated.}
    \label{fig:sam3-vlm-teaser}
\end{figure}

\begin{figure*}[!htpb]
\centering
\includegraphics[width=\linewidth]{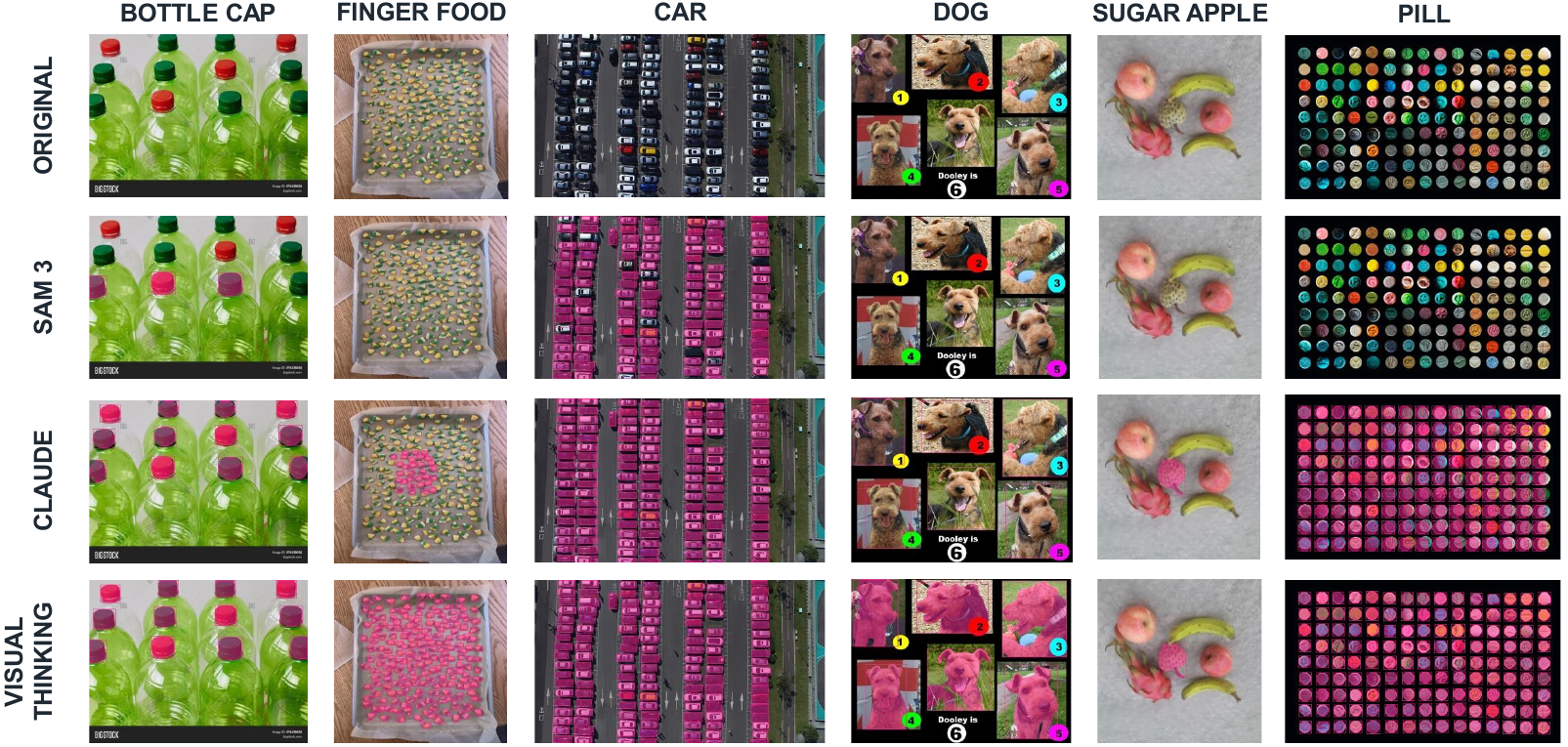}
\caption{Examples of Visual Thinking improving counting and localization compared to SAM 3 and Claude Opus 5.5 (with extra-high reasoning effort) on FSCD-147, CARPK, PixMo-Count, and OmniCount (Fruits) datasets.}
\label{fig:datasets}
\end{figure*}

Recent developments in vision foundation models (VFMs) have made zero-shot object detection possible across diverse images and object categories. Models such as SAM 3 \cite{sam3} and Grounded SAM 2 \cite{groundedsam2} support text-guided detection and segmentation of the objects without additional task-specific fine-tuning. Despite their strong performance for clearly visible or standard objects, small, occluded and unusual objects remain challenging. Detailed text prompts can introduce further difficulties. SAM 3’s concept-prompt interface, for example, is designed around short noun phrases, while identifying the intended objects may require interpreting diverse attributes, and scene context.

Vision-language models (VLMs) can help bridge this gap by interpreting detailed instructions and translating them into descriptions of visible objects. However, translating this understanding into precise bounding boxes or segmentation masks remains challenging. Combining these models, seems like a beneficial solution: a VLM interprets the target and guides the search, while a VFM provides precise object localization. Existing systems, including SAM 3 Agent~\cite{sam3} and the LLM-controlled refinement procedures in CountGD++~\cite{countgdpp}, demonstrate the potential of this collaboration.

Finding every instance of a target, however, requires deciding how to search further and what to retain from each search. A VLM may suggest descriptions that help recover objects missed by the original prompt. For example, ``green fruit'' may reveal citruses missed by ``citrus'', but it may also return green apples or citrus fruits already detected. Such results require different treatment: new targets should be added, repeated detections should be matched to existing objects, and unrelated objects should be excluded. Later observations can also provide evidence for reconsidering earlier detections. The search therefore needs to track individual objects and the evidence collected about them across prompts.

To address these, we introduce \textbf{Visual Thinking}, a training-free framework that connects a VLM and SAM 3 through this iterative perception loop. Starting from detections obtained with the original target prompt, the VLM examines the image and current results to propose alternative descriptions. These descriptions can emphasize different attributes or vary in semantic breadth. A controller orders the proposed searches using their relevance to the target, uncertainty about existing candidates, and previously untested sensing conditions. Automatically selected positive visual exemplars and spatial refinement provide further guidance to SAM 3. Across searches, Visual Thinking matches repeated detections to the same object and combines their SAM 3 scores using a weighted average. Narrow target descriptions receive greater weight than broad descriptions, allowing different prompts to contribute evidence without treating their scores equally. Newly discovered objects are added to the object library, while further observations can update confidence in existing detections. The VLM receives these updated results and the search history to guide its next proposals. We evaluate counting and localization on CARPK, FSCD-147, PixMo-Count, and OmniCount (Fruits). Compared with SAM 3, Visual Thinking reduces the counting mean absolute error (MAE) by about 18 on FSC-147 and about 8 on CARPK. It improves box average precision (AP) by more than 12 on FSCD-147 and CARPK. Visual Thinking, with fixed or no tiling, performs on par with Claude Opus 5.5 at extra-high reasoning effort, while running on single desktop GPU in about 5.91 seconds per image, compared to about 46.53 seconds per image for Claude through its API.

\begin{figure*}
\centering
\includegraphics[width=\linewidth]{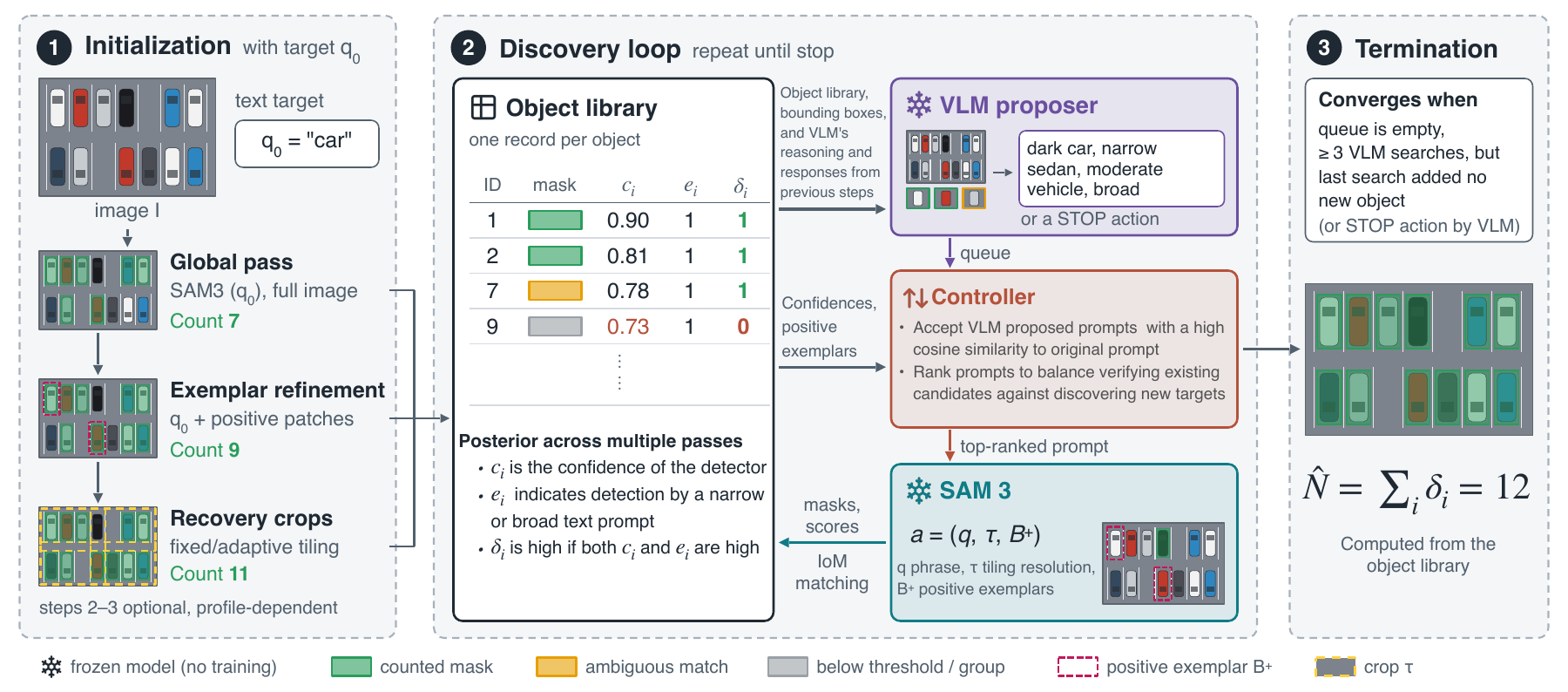}
\caption{\textbf{Our approach for visual thinking for object detection has three main parts.}
(1) \textbf{Initialization.} We query SAM 3 with the target prompt $q_0$ on the full image, then use positive exemplars to increases recall and, depending upon the dataset, perform a third query on crops of the image to recover small objects. These detections are used to initialize an object library with one entry per object that consists of a mask $m_i$, a confidence $c_i$ (weighted mean of SAM 3 scores) and an indicator $e_i$ of whether the object was detected confidently by a semantically narrow phrase.
(2) \textbf{VLM proposer.} A frozen VLM is provided the image, crops of objects in the library, the VLMs actions from past iterations and the outcomes of its past actions. It proposes up to three short phrases, with each phrase labeled as being semantically narrow, moderate, or broad relative to the original prompt $q_0$.
(3) \textbf{Controller.} The controller discards phrases unrelated to $q_0$ and ranks the rest to achieve a balance between improving the uncertainty of existing object detections and discovering new objects using different text prompts. SAM 3 is used to execute the top-ranked action $a = (q, \tau, B)$, with phrase $q$, crops $\tau$ and exemplars $B$. Detections from SAM 3 are matched to the library using intersection-over-minimum metric (IoM) and confidence estimates of objects in the library are updated at each step.
This system has no trainable parameters.
}
\label{fig:sam3-vlm-pipeline}
\end{figure*}

\section{Related Work}
\label{s:related}
\noindent\textbf{Open-vocabulary detection and segmentation.} Open-vocabulary detectors allow target objects to be localized via natural language descriptions or visual exemplars. Pioneering methods like OWL-ViT~\cite{owlvit} adapt standard Vision Transformers~\cite{vit} directly for open-vocabulary detection, while Grounding DINO~\cite{groundingdino} extends the DINO detector~\cite{dino} via grounded pre-training and deep cross-modal feature fusion. In parallel, promptable segmentation foundation models such as SAM~\cite{sam} and SAM~2~\cite{sam2} generate precise masks from spatial prompts like points and bounding boxes. Building on these foundations, Grounded SAM~2~\cite{groundedsam2}
connects text-guided detection with mask extraction, while SAM~3~\cite{sam3} accepts text and visual exemplars to detect and segment matching instances.

\noindent\textbf{Text- and exemplar-guided counting.} Text- and exemplar-guided counting methods count target objects in an image based on a text prompt or visual example. Early methods like CounTX~\cite{countx} and CLIP-Count~\cite{clipcount} focus on counting via text descriptions, while CountGD~\cite{countgd} supports both text prompts and visual exemplars. To remove the need for manual exemplar annotation, subsequent works automate exemplar selection: CountSE~\cite{countse} extracts soft exemplars directly from image features, while CountGD++~\cite{countgdpp} selects pseudo-exemplars from initial model detections and supports LLM-controlled iterative refinement. In addition, recent approaches explore spatial search strategies to handle difficult images.  SAM3Count~\cite{sam3count} uses adaptive tiling, while AdaCount~\cite{adacount} utilizes prototype similarity to guide spatial warping and feature modulation without additional training.

\noindent\textbf{Sequential visual reasoning and search.} Sequential visual reasoning frameworks query models iteratively, using intermediate predictions to guide subsequent steps and refine the final output. For example, Variational Information Pursuit~\cite{vip} sequentially selects queries to produce informative and interpretable predictions, with an extension using large language models~\cite{vip-llm} to automate query generation. Similarly, agent-based methods refine visual search over multiple rounds: SAM 3 Agent~\cite{sam3} inspects intermediate segmentation outputs to update its prompts for subsequent queries, while VASA~\cite{vasa} maintains a working mask to track and accumulate results across reasoning steps.

\section{Method}
\label{s:method}

Given an image $I$ and a text query $q_0$ for the category of the object to be detected, we want to count the number of instances of that object and segment each of these instances. The integral components of our system are a vision-language model (VLM) and an object detector like SAM 3 that takes a text query and produces bounding boxes or segmentation masks for different instances that are semantically related to the query. We can think of the interaction between the VLM and the detector as an active perception problem \cite{bajcsy1988active}. At each iteration, the system selects an action $a = (q, \t, B)$ that consists of a short noun phrase $q$, a set $\t$ potentially overlapping crops that cover the image $I$ and a set $B$ of positive exemplar crop. SAM 3 is the ``sensor'' that runs on each crop with the text query $q$ and exemplars $B$. Suppose the sensor returns a set of masks and scores denoted by $y$. The active perception problem involves selecting the next action that maximizes the information gain
\beq{
     a_{t+1} \in \operatorname{argmax}_a \text{I}(Y_{t+1};\ y_1, \dots, y_t \mid a).
     \label{eq:ig}
}
where $\text{I}$ denotes the mutual information. It can be shown that instantiating this procedure requires (i) a representation of past observations that can be updated incrementally using the new observations, and (ii) a criterion to choose among putative actions \cite{siming2024active}. This recipe works well for problems such as selecting the next viewpoint for a robot to explore an unknown region, where a map of the environment built using, say, a neural radiance field acts as the representation and candidate actions consist of putative viewpoints. The mutual information between future observations and the current map (which is a representation of the past observations) can be computed for different viewpoints using the rendering equation. This recipe is difficult to instantiate in our setting because models like VLMs do not always allow access to the likelihood.
We next describe the main components of our approach depicted in \cref{fig:sam3-vlm-pipeline}.

\paragraph{Object library}
The representation of past observations in our problem is an object library $L_t = \cbr{(m_i, c_i, e_i)}_{i=1}^{M_t}$ which contains a mask $m_i$, a confidence $c_i$ and an indicator $e_i$ that denotes whether object $i$ was detected confidently by a semantically narrow prompt (more details below). We will compare masks using the intersection over minimum
\[
     \text{IoM}(m, m') = \f{\abs{m \cap m'}}{\min(\abs{m}, \abs{m'})}
\]
which is large when one mask is a fragment of the other, e.g., an object that is cut by a crop boundary. We suppress a detection if its IoM with a higher-scoring detection for this action is at least a pre-specified threshold $\k$. Each detection either gets associated to an existing entry in the object library if their IoM is larger than $\k$, or spawns a new entry. We imagine visual thinking as iteratively selecting prompts based on observations from SAM 3. Detections are precise for semantically narrow prompts while semantically broad prompts enable a higher recall. The VLM therefore provides each text-query a ``semantic breadth'' $b(q) \in \{\text{narrow, moderate, broad}\}$ relative to the original query $q_0$ (which is defined to have a narrow breadth). Let $s_{ik}$ be the score of object $i$ for action $a_k$ and let $A_i$ be the actions that detected object $i$ in the library without using it as an exemplar. The confidence of object $i$ is the weighted mean
\[
     c_i = \f{\sum_{k \in A_i} w_k s_{ik}}{\sum_{k \in A_i} w_k}
\]
where $w_k$ depends upon the breadth of the phrase $q_k$ with $w_{\text{narrow}} \geq w_{\text{moderate}} \geq w_{\text{broad}} > 0$. Broad phrases have small weights because they can also detect visually similar objects that are not the target object. The confidence of an object can also go down if the score $s_{ik}$ is low for a chosen action. Therefore, it is not sufficient to detect objects using high-scoring broad actions, we also require that objects be detected confidently using narrow phrases,
\beq{
     e_i = \mathbf{1}\cbr{\exists k \in A_i: b(q_k) = \text{narrow}, s_{ik} \geq \theta},
     \label{eq:ei}
}
where $\theta$ is a pre-specified threshold. Objects that have high confidences in the object library and have been detected using narrow phrases are considered valid, i.e.,
\beq{
     \hat N_t = \sum_{i=1}^{M_t} \delta_i,\quad \text{where } \delta_i = \mathbf{1}\cbr{c_i \geq \th} e_i.
     \label{eq:hatn}
}
High-confidence objects $i$ with $\delta_i = 1$ in the object library act as positive exemplars $B_t$ for the next iteration.

\paragraph{Initializing the object library by iteratively querying SAM 3}
The first query runs globally over the whole image with the original text prompt without any exemplars, i.e., $a_0 = (q_0, \{I\}, \emptyset)$. This helps initialize the object library by providing exemplars $B_1$ to run the next action globally, i.e., $a_1 = (q_0, \{I\}, B_1)$ before action $a_2 = (q_0, \t, B_2)$ tiles the image to recover small objects. The tiling $\t$ is either a fixed grid, or it is computed from the current library. The crop size is proportional to the typical size of detected objects. Regions with many small objects are split into smaller crops, and the rest of the image is covered by larger ones, similar to SAM3Count \cite{sam3count}.

\paragraph{Discovery loop and the VLM proposer}
In the discovery loop, the VLM proposes new phrases, see the prompt in \cref{fig:qwen_prompt}. At each iteration, the VLM receives an image, coordinates of bounding boxes for objects in the library sampled by their confidences, the actions that were executed in previous iterations and their outcomes, along with previous VLM responses. The VLM returns up to three pairs $(q, b(q))$ with text prompt $q$ having up to three words. The VLM can also return a STOP action.  After lemmatization of the prompts $q$, phrases that were already executed or have been already queued for SAM 3 are rejected. The remaining phrases are passed to a controller module described next.

\paragraph{Controller}
The controller runs SAM 3 on phrases from the queue one at a time, and produces observations $y_{t+1}$ to update the library to $L_{t+1}$.
It queries the VLM if the queue is empty. We say that an action discovers an object $i$ in the library if it results in changing $\delta_i = 1$ from zero for the first time.
The controller terminates the active perception procedure if the queue is empty and at least $k = 3$ VLM phrases have been processed but the last phrase did not discover any object.

At each iteration, the controller performs a preliminary filtering of the queue using rescaled cosine similarity $R(q)$ of the normalized text embedding (computed using a sentence-transformers model named all-MiniLM-L6-v2~\cite{all-MiniLM-L6-v2}) to reject phrases proposed by the VLM that are dissimilar to the target phrase. The controller then ranks the queue by balancing the verification of existing objects against the discovery of new ones. Object $i$ is considered ``unresolved'' if the latest detection associated with it (i.e., with $\text{IoM} \geq \k$) also has IoM at least $\k$ with some other object in the library, or if $c_i \geq \theta$ but $e_i = 0$. We set the verification priority of object $i$ to be
\[
u_i = \max\cbr{\text{H}(c_i), \mathbf{1}\{i \text{ is unresolved}\}},
\]
where $\text{H}$ is the base-2 binary entropy. Let $V_t$ be the set of objects in the object library that are unresolved or whose confidence $\abs{c_i - \th} \leq \e$ for a threshold $\e = 0.05$. Let
\[
U_t = \frac{1}{\max(1, \abs{V_t})} \sum_{i \in V_t} u_i
\]
be their average priority.

Discovering new objects requires the VLM to use new text phrases for querying SAM 3. To enable this, we maintain descriptors of text prompts $q$, e.g., words in the prompt beyond the main noun (``small'' in ``small car''). Each action $a_t = (q_t, \t_t, B_t)$ is tagged with these words, the indicator variable $\mathbf{1}\{B_t \neq \emptyset\}$ and a scalar that is the median of the logarithm of the longer side of crops in $\t_t$ (converted to an integer). Let this tag be $F(q)$. If $F_t$ is the union of tags across actions $a_0, a_1, \dots, a_t$, then the ``novelty'' of a prompt $q$ is the fraction of its tags that are not yet evaluated, i.e., $D_t(q) = \abs{F(q) \setminus F_t}/\abs{F(q)}$. All phrases in the queue $Q_t$ share $\tau_t$ and $B_t$, so their novelty differs only through their descriptors. The next action selected by the controller is $a_{t+1} = (q^*, \tau_t, B_t)$, where
\beq{
q^* = \arg\max_{q \in Q_t} R(q) \sbr{\alpha U_t + (1 - \alpha) D_t(q)}.
\label{eq:qstar}
}
A larger $\alpha \in (0, 1)$ favors verifying existing objects, and a smaller one favors new prompts. This score is a heuristic alternative for the information gain in \cref{eq:ig}.

\paragraph{Using frontier models for visual thinking}
Frontier large language model-based systems that can operate on both images and text using reasoning can also be prompted to perform visual thinking. We used Claude Opus 5.5 (with extra-high reasoning) for this purpose. One of our key considerations was to prevent the model from looking up answers to the task on the Internet, call existing detection-models (like SAM 3, which would make it equivalent to our proposed approach) using tool calls or build local variants of such detection models itself, prior published results, or leader-board scores. This required an elaborate prompt that was crafted by trial-and-error, which is shown in \cref{fig:claude_prompt}. Our workflow with Claude has two major steps. First, it inspects the image using operations such as zoom, tiling, cropping with resize and rotations disallowed (to keep results consistent while estimating metrics). Second, it produces a JSON entry that contains instances, bounding boxes, a visual text-based summary of the image, a text-based explanation for the outcome, a method summary that describes the approach that was used, operations that were performed upon the image, and a textual description of the uncertainty. Claude runs one-shot with extra-high reasoning on each image independently.

We found that bounding boxes predicted by Claude are largely accurate, although not as accurate as those of SAM 3. Segmentation masks predicted by Claude are quite poor and they are largely elliptical or rectangular in shape regardless of the shape of the underlying object. Nevertheless, as we will discuss next, Claude performs remarkably well on benchmark tasks by utilizing tricks like estimating object counts in busy images by extrapolation from counts within specific patches.

\section{Experiments}
\label{s:results}

We evaluate Visual Thinking on four benchmark datasets covering different object categories, densities, and types of scenes. We also discuss results on a new dataset that consists of particularly challenging examples of object localization and counting. \cref{fig:datasets} (top row) shows some example images from the datasets used in our analysis. \cref{s:app:implementation_details} discusses implementation details of the Visual Thinking approach.  In ablation experiments in \cref{s:app:ablation_experiments}, we compare global detection, exemplar and spatial refinement, and VLM-guided search to identify where the improvements come from.

\begin{figure*}[!htpb]
\centering
\includegraphics[width=\linewidth]{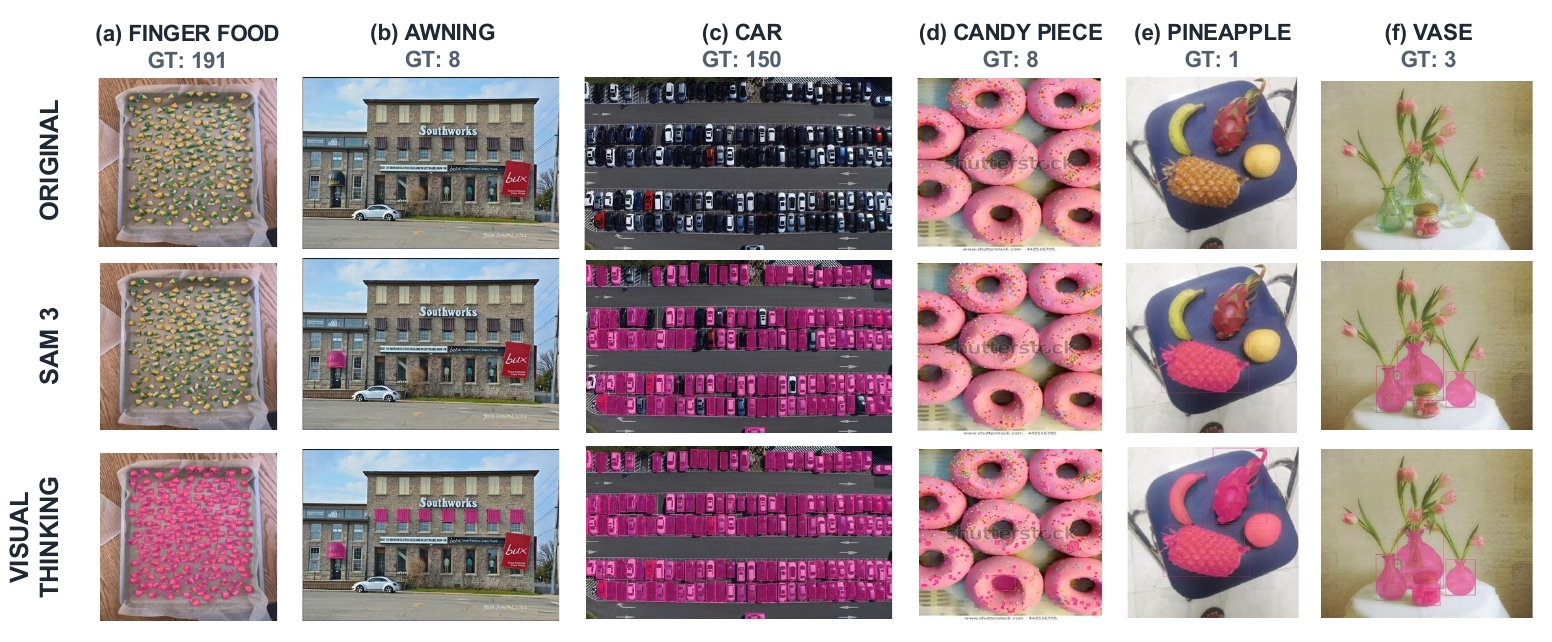}
\caption{\textbf{Successes and failures of Visual Thinking.} (a) The abstract category ``finger food'' recovers no objects in the SAM 3 pass. The VLM's rewordings into ``corn candy'' and ``triangular candy'' recover every instance. (b) Added descriptors of the structure and material (``window shade structure'' and ``striped fabric cover'') find awnings initially missed by SAM 3. (c) The rewording into ``vehicle'' and positive exemplars help recover the missing cars. (d) A mislabeled original query steers the rewording towards ``candy sprinkle'' recovering unintended objects. (e) The VLM's rewording drifts into broader categories such as ``spiky fruit'' which adds dragon fruit, and ``textured oval fruit'', which recovers every fruit in the image. (f) The VLM proposes the broad prompt ``jar'' which recovers a small jar in front of the vases.}
\label{fig:success-failures}
\end{figure*}

\paragraph{Baselines}
We compare with four groups of methods.
The first group, for FSC-147, consist of methods like T2ICount~\cite{t2icount} and QICA~\cite{qica} which regress a density map. PseCo~\cite{pseco} prompts SAM with a set of points and classifies the resulting masks. DAVE~\cite{dave} keeps clusters with a high CLIP similarity to the text prompt and then removes outliers in a verification step. CountGD~\cite{countgd} extends Grounding DINO~\cite{groundingdino} to accept text and visual exemplars and predicts points, while CountGD-Box~\cite{countvid} extends this approach to output bounding boxes in addition to points. CountGD++~\cite{countgdpp} takes the top few confident detections as pseudo exemplars while CountSE~\cite{countse} can also use soft exemplars selected from image features at multiple scales.
The second group consists of training-free methods built on foundation models. OmniCount~\cite{omnicount} uses semantic and geometric priors from pretrained models to prompt SAM. SAM3Count~\cite{sam3count} can run SAM 3 on adaptively sized tiles. AdaCount~\cite{adacount} runs SAM 3 twice. Features are pooled from confident detections to get a similarity map of target regions. The second pass enlarge these regions, amplifies their SAM 3 features, and maps the detections back to the original image.
The third group is segmentation models such as SAM~3~\cite{sam3} prompted with the target description on the full image. This is equivalent to the first action $a_0$ of our method, so it also acts as an ablation study.
The fourth group is a frontier model, Claude Opus 5.5~\cite{claudeopus55}, prompted as discussed in \cref{fig:claude_prompt}. It inspects the image by zooming, tiling and cropping, and returns a box for each instance.

\paragraph{Metrics}
For counting tasks, we report the mean absolute error (MAE), i.e., the difference between the predicted count and the ground-truth count. We also report the root mean squared error (RMSE) which enables distinguishing between methods that have few images with a large MAE versus many images with small MAE. For some datasets like PixMo-Count, we also report the accuracy of the counts, i.e., the fraction of images where a counting method predicts the ground-truth count exactly. For localization tasks, we report average precision (AP) which is calculated across multiple Intersection over Union (IoU) thresholds from 0.50 to 0.95 with steps of 0.05. We also report the average precision at an IoU threshold of 0.5 (AP50).

\subsection{Results} 
\label{s:r1}

\begin{table}[!htpb]
\centering
\small
\caption{
Object counting results on the test split of the FSC-147 dataset.
Train denotes methods that were trained on this task explicitly with access to the training dataset.
The second block shows zero-shot metrics for different methods on this task, as reported in the original papers.
The third block shows zero-shot metrics obtained in our experiments.
}
\label{tab:fscd147-counting}

\setlength{\tabcolsep}{4pt}
\renewcommand{\arraystretch}{1.05}

\begin{adjustbox}{width=\linewidth}
\begin{tabular*}{\columnwidth}{
    @{\extracolsep{\fill}}lcrrr@{}
}
\toprule[1pt]
Method & Train & MAE $\downarrow$ & RMSE $\downarrow$ & Runtime (s) \\
\midrule
PseCo~\cite{pseco}   & \cmark    & 16.58 & 129.77 & ---\\
DAVE~\cite{dave}    & \cmark       & 14.90 & 103.42  & ---\\
CountSE~\cite{countse}   & \cmark        & 7.84 & 82.99 & --- \\
QICA~\cite{qica}      & \cmark     & 12.41 & 97.28 & --- \\
T2ICount~\cite{t2icount}   & \cmark    & 11.76 & 97.86  & ---\\                
CountGD~\cite{countgd}  & \cmark & 12.98 & 98.35 & --- \\
CountGD++~\cite{countgdpp}  & \cmark    & 16.55 & 129.76 & --- \\           
\midrule
SAM~\cite{sam}     & \xmark       & 42.48 & 137.50 & --- \\
OmniCount~\cite{omnicount}  & \xmark    & 21.46 & 133.28 & --- \\
SAM3Count~\cite{sam3count}   & \xmark   & 19.80 
                          & 128.59 & --- \\
AdaCount~\cite{adacount}   & \xmark    & 18.97
                          & 126.94 & --- \\
\midrule
SAM 3~\cite{sam3} & \xmark
     & 33.47 & 149.62 & 0.16\\
Opus 5.5 (est.)~\cite{claudeopus55} & \textbf{?}
     & 10.50 & 44.53 & 53.85\\
Opus 5.5 (det.)~\cite{claudeopus55} & \textbf{?}
     & 30.07 & 148.72 & 53.85 \\
Visual Thinking (ours) & \xmark
     & 14.78 & 129.37 & 50.95 \\
\bottomrule[1pt]
\end{tabular*}
\end{adjustbox}
\end{table}

\begin{table}[!htpb]
\centering
\small
\caption{
Object localization on bounding boxes on test split of the FSCD-147 dataset.
AP indicates average precision (AP) across multiple Intersection over Union (IoU) thresholds, and AP50 is average precision at an IoU threshold of 0.5.
}
\label{tab:fscd147-localization}
\begin{adjustbox}{width=0.9\linewidth}
\renewcommand{\arraystretch}{1.05}
\begin{tabular*}{\columnwidth}{@{\extracolsep{\fill}}lrr@{}}
\toprule[1pt]
Method & AP $\uparrow$ & AP50 $\uparrow$ \\
\midrule
PseCo~\cite{pseco}
    & 37.91 & 62.45 \\
DAVE~\cite{dave}
    & 18.50 & 50.24  \\
CountGD~\cite{countgd}
    & 18.10 & 52.90  \\
CountGD-Box~\cite{countvid}
    & 30.44 & 61.56  \\
CountGD++~\cite{countgdpp}
    & 38.93 & 71.35  \\
SAM 3~\cite{sam3}
    & 32.24 & 48.83  \\
Visual Thinking (ours)
    & 44.28 & 72.30 \\
\bottomrule[1pt]
\end{tabular*}
\end{adjustbox}
\end{table}

\paragraph{FSC-147}~\cite{fsc147} is a counting dataset with 147 classes and 6,135 images with objects counts per image ranging from as low as 7 to as high as 3,731. Each test datum consists of an image and the desired object class in natural language, which are both used as input to our system. Due to the large number of objects, the original paper that developed the dataset focused on detection methods that predict a density map for objects of interest. This dataset has a wide spectrum of scenes (sparse to dense) and object types (small ones to large ones), so it acts as a broad, generalist benchmark for our approach.

\cref{tab:fscd147-counting} reports results on the FSC-147 dataset alongside comparisons to several baseline approaches.
Observe that methods like CountSE~\cite{countse} and T2ICount~\cite{t2icount}, that are based on supervised-learning using the training data achieve MAEs ranging from 7.84--11.76.
SAM 3~\cite{sam3}, which was not trained on this data, achieves a zero-shot MAE of 33.47.
Visual Thinking reduces the MAE from 33.47 to 14.78, and the RMSE from 149.62 to 129.37, compared to SAM 3.
Claude Opus 5.5~\cite{claudeopus55}, when prompted to perform this task as we discussed above, achieves an MAE of 10.50. This is remarkably high. And in fact, we noticed that for images in this dataset, Claude estimates the counts based on detections inside a small number of patches---rather than explicitly detecting each object. This is an interesting demonstration of ``visual thinking'', very much along the lines of our work. If we evaluate Claude in terms of its explicit detections, the MAE is much higher 30.07, and not substantially better than SAM 3's.
Among all zero-shot methods, our approach achieves the second lowest MAE, after Claude.

\paragraph{FSCD-147}~\cite{fscd147} is an extension of the FSC-147 dataset that introduces bounding-box annotations for each object. We evaluate counting and localization on 1,286 validation images and 1,190 test images. We use the singularized version of the object class as the object category for the initial prompt. This benchmark evaluates detection and localization performance across different object categories, sizes, and densities.

For the localization task, we compare Visual Thinking with existing methods in \cref{tab:fscd147-localization}. Visual Thinking outperforms existing methods substantially in terms of both AP and AP50. We can understand this as follows. In \cref{tab:fscd147-counting}, SAM 3 is substantially worse than supervised methods, say, CountGD, which suggests that it has a poor recall. But SAM 3 is much better at object localization compared to these methods, as seen in the original paper \cite{sam3}. Visual Thinking achieves a much better AP because it can iteratively detect more objects compared to SAM 3 and, because it uses SAM 3, it can also localize these objects more accurately than existing localization methods. Visual Thinking therefore achieves the best of both worlds---good precision and good recall.
Claude does not provide estimates of the confidence score for detections, and therefore we cannot calculate AP.

\begin{table}
\centering
\small
\caption{
Zero-shot evaluation of object counting for different existing methods on the test split of the CARPK dataset.
}
\label{tab:carpk-counting}

\renewcommand{\arraystretch}{1.05}
\begin{adjustbox}{width=0.9\linewidth}
\begin{tabular*}{\linewidth}{
    @{\extracolsep{\fill}}lrrr@{}
}
\toprule[1pt]
Method & MAE $\downarrow$ & RMSE $\downarrow$ & Runtime (s) $\downarrow$\\
\midrule

CountSE~\cite{countse}           & 2.79 & 4.20 & --- \\
QICA~\cite{qica}           & 6.07 & 7.82 & --- \\
T2ICount~\cite{t2icount}       & 8.61 & 13.47 & --- \\                
CountGD~\cite{countgd}  & 3.83 & 5.41 & --- \\
CountGD++~\cite{countgdpp}      & 2.48 & 3.74 & --- \\           
OmniCount~\cite{omnicount}      & 13.41 & 16.85  & --- \\
SAM3Count~\cite{sam3count}      & 3.11
                          & 5.60  & --- \\
AdaCount~\cite{adacount}       & 1.99
                          & 3.05 & --- \\
Claude Opus 5.5~\cite{claudeopus55}
     & 1.92 & 2.76 & 161.66 \\
SAM 3~\cite{sam3}
     & 9.50 & 13.44 & 0.16 \\
SAM 3 + four tiles~\cite{sam3}
     & 1.58 & 2.69 & 0.82 \\
Visual Thinking (ours)
     & 1.47 & 2.60 & 8.26 \\

\bottomrule[1pt]
\end{tabular*}
\end{adjustbox}
\end{table}

\paragraph{CARPK}~\cite{carpk} is a vehicle counting dataset with 459 aerial drone images of vehicles in a parking lot with 2--188 cars per image in its test split. Each test datum consists of the image and the text-prompt which is simply ``car'' for all images. We use it to to assess counting and localization of repeated, closely spaced objects.

Many existing approaches have used this dataset for zero-shot evaluation.
For all methods, metrics in \cref{tab:carpk-counting,tab:carpk-localization} are much better compared to those in \cref{tab:fscd147-counting,tab:fscd147-localization} because the semantic ambiguity of objects in CARPK is much smaller than FSCD-147. Visual Thinking is significantly better than existing methods, including Claude, for both object detection and localization. Images in CARPK typically contain a large number of cars. Therefore, we found it useful to tile the image into four quadrants and run SAM 3 and our system independently on the four tiles. This step was inspired from how Claude used tiling to handle these images. As we see in \cref{tab:carpk-counting}, the accuracy of tiling is much better than non-tiled methods like vanilla SAM 3. This benchmark demonstrates that our system is effective for medium-density scenes with lots of repeated instances of similar objects.

\begin{table}[!htpb]
\centering
\small
\caption{
Object localization on bounding boxes on test split of the CARPK dataset.
AP indicates average precision (AP) across multiple Intersection over Union (IoU) thresholds, and AP50 is average precision at an IoU threshold of 0.5.
}
\label{tab:carpk-localization}

\setlength{\tabcolsep}{4pt}
\renewcommand{\arraystretch}{1.05}
\begin{adjustbox}{width=0.9\linewidth}
\begin{tabular*}{\linewidth}{@{\extracolsep{\fill}}lrr@{}}
\toprule[1pt]
Method & AP $\uparrow$ & AP50 $\uparrow$ \\
\midrule
SAM 3~\cite{sam3}
    & 52.05 & 79.36 \\
SAM 3 + four tiles~\cite{sam3}
    & 64.43 & 93.88 \\
Visual Thinking (ours)
    & 64.53 & 93.88 \\
\bottomrule[1pt]
\end{tabular*}
\end{adjustbox}
\end{table}

\begin{table}[!htpb]
\centering
\small
\caption{
Zero-shot evaluation of object counting for different existing methods on the test split of the PixMo-Count dataset.
}
\label{tab:pixmocount-counting}

\setlength{\tabcolsep}{4pt}
\renewcommand{\arraystretch}{1.05}

\begin{adjustbox}{width=0.9\linewidth}
\begin{tabular*}{\linewidth}{
    @{\extracolsep{\fill}}lrrr@{}
}
\toprule[1pt]
Method & MAE $\downarrow$ & Acc (\%) $\uparrow$ & Runtime (s) $\downarrow$ \\
\midrule
SAM3Count~\cite{sam3count} & 0.17 & 89.18 & --- \\
Claude Opus 5.5~\cite{claudeopus55}
     & 0.30 & 83.37 & 16.36 \\
SAM 3~\cite{sam3}
     & 0.23 & 87.19 & 0.16 \\
Visual Thinking (ours)
     & 0.15 & 89.10 & 5.36 \\

\bottomrule[1pt]
\end{tabular*}
\end{adjustbox}
\end{table}

\paragraph{PixMo-Count}~\cite{molmo} consists of images with counts ranging from 2 to 10, testing counting in scenes with relatively few target instances. The challenges in this dataset come from occluding objects, multiple instances of the same semantic category, and quite different viewpoints across images. Since the dataset does not provide bounding boxes annotation, we report counting metrics only. This dataset also contains an image and a text query for the desired object, which we singularize before using it in our system.

Since the number of objects in each image is much smaller than other benchmark datasets, the MAE of all methods, including Visual Thinking in \cref{tab:pixmocount-counting} is as low as 0.15. The accuracy, which measures whether the object count was exactly correct, is also rather high. It is also interesting to note that Visual Thinking without tiling performs as well as SAM3Count \cite{sam3count} which uses adaptive region-of-interest tiling.

\begin{table}[!htpb]
\centering
\small
\caption{
Counting results on the test split of OmniCount (Fruits).
}
\label{tab:omnicount-counting}
\renewcommand{\arraystretch}{1.05}
\begin{adjustbox}{width=0.9\linewidth}
\begin{tabular*}{\linewidth}{
    @{\extracolsep{\fill}}lrrr@{}
}
\toprule[1pt]
Method & MAE $\downarrow$ & RMSE $\downarrow$ & Runtime (s) $\downarrow$ \\
\midrule   
CountGD~\cite{countgd}  & 2.76 & 3.11 & ---\\
CountGD++~\cite{countgdpp}     & 1.97 & 3.29 & --- \\           
SAM3Count~\cite{sam3count}      & 0.43 & 0.93 & ---\\
AdaCount~\cite{adacount}       & 0.37
                          & 0.74  & ---\\
Claude Opus 5.5~\cite{claudeopus55}
     & 0.02 & 0.12 & 15.80\\
SAM 3~\cite{sam3}
     & 0.34 & 0.60 & 0.16 \\
Visual Thinking (ours)
     & 0.33 &0.63 & 4.74 \\

\bottomrule[1pt]
\end{tabular*}
\end{adjustbox}
\end{table}

\begin{table}[!htpb]
\centering
\small
\caption{
Object localization on bounding boxes on test split of the OmniCount (Fruits) dataset.
}
\label{tab:omnicount-localization}

\renewcommand{\arraystretch}{1.05}

\begin{adjustbox}{width=0.9\linewidth}
\begin{tabular*}{\linewidth}{@{\extracolsep{\fill}}lrr@{}}
\toprule[1pt]
Method & AP $\uparrow$ & AP50 $\uparrow$ \\
\midrule
CountGD-Box~\cite{countvid}
    & 47.00 & 61.00 \\
CountGD++~\cite{countgdpp}
    & 54.00 & 72.00 \\
Claude Opus 5.5~\cite{claudeopus55}
    & 58.76 & 90.08 \\
Visual Thinking (ours)
    & 41.20 & 64.22 \\
\bottomrule[1pt]
\end{tabular*}
\end{adjustbox}
\end{table}

\paragraph{OmniCount (Fruits)}~\cite{omnicount} is a subset of the OmniCount benchmark containing 303 images with 8 fruit classes. This benchmark tests whether the pipeline can count and localize a specified fruit category among other fruits. It is challenging for two reasons: test images contain fruits of different types while the text prompt may only select a few specific fruits. Methods that achieve good performance in this benchmark are those that can detect and localize the fruit within the image while using the text-prompt in precise fashion to avoid detecting other fruits.

As we see in the previous examples, the zero-shot MAE of Visual Thinking on object counting \cref{tab:omnicount-counting} is very good. On the other hand, the localization performance in \cref{tab:omnicount-localization} is worse than existing methods like CountGD++. In both cases, we observed that the performance degrades because SAM 3 does not recognize the category ``sweet apple'', and often mistakes it to be the co-occurring red apple. Claude performs significantly well on this benchmark in both tasks. This experiment shows that Visual Thinking can improve performance beyond SAM 3 (as seen in the other benchmarks), but its performance depends upon the vocabulary used to train the underlying object detection model.

\paragraph{Green Citruses Dataset} is a new dataset that consists of images of green citruses on green foliage. The test split has 35 images and the train split as 631 images. We created the ground-truth bounding boxes for the test split, which is used in our analysis, ourselves.
This is a very challenging dataset, and in spite of the fact that citruses occupy a substantial number of pixels, they are very difficult to detect against the background---even for humans. There is significant lighting variation across images because they were collected in an outdoor environment using relatively low-resolution cameras. Such data is quite typical in real-world computer vision applications. We use this dataset to show that our system successfully recovers very hard-to-find objects in highly occluded environments.

The green citruses in this benchmark are very small. Therefore, as in our CARPK evaluation, we divide the image into nine tiles and run SAM 3 and Visual Thinking on each of the tiles independently.  As shown in \cref{tab:greencitrus-counting}, tiling alone slightly reduces MAE compared with vanilla SAM 3. Visual Thinking achieves a larger improvement, reducing MAE from 26.26 to 14.14. It also achieves lower MAE than Claude, highlighting the effectiveness of our approach for detecting small, occluded objects in real-world settings.

\begin{table}[!htpb]
\centering
\small
\caption{
Object counting on test split of the Green Citruses dataset.
}
\label{tab:greencitrus-counting}
\renewcommand{\arraystretch}{1.05}
\begin{adjustbox}{width=0.9\linewidth}
\begin{tabular*}{\linewidth}{@{\extracolsep{\fill}}lrrr@{}}
\toprule[1pt]
Method & MAE $\downarrow$ & RMSE $\downarrow$ & Runtime (s) $\downarrow$\\
\midrule
SAM 3~\cite{sam3} & 26.26 & 28.18 & 0.16 \\
SAM 3 + nine tiles~\cite{sam3} & 24.00 & 25.79 & 1.60 \\
Claude Opus 5.5~\cite{claudeopus55} & 16.20 & 18.49 & 63.60 \\
Visual Thinking (ours) & 14.14 & 18.33 & 23.96\\
\bottomrule[1pt]
\end{tabular*}
\end{adjustbox}
\end{table}

\begin{table}[!htpb]
\centering
\small
\caption{
Object localization on bounding boxes on test split of the Green Citruses dataset.
}
\label{tab:citruses-localization}

\setlength{\tabcolsep}{4pt}
\renewcommand{\arraystretch}{1.05}
\begin{adjustbox}{width=0.9\linewidth}
\begin{tabular*}{\linewidth}{@{\extracolsep{\fill}}lrr@{}}
\toprule[1pt]
Method & AP $\uparrow$ & AP50 $\uparrow$ \\
\midrule
SAM 3~\cite{sam3}
    & 2.34 & 4.85 \\
SAM 3 + nine tiles~\cite{sam3}
    & 5.35 & 10.12 \\
Visual Thinking (ours)
    & 19.65 & 36.87 \\
\bottomrule[1pt]
\end{tabular*}
\end{adjustbox}
\end{table}

\paragraph{Runtime} We report mean runtime per image in  \cref{tab:fscd147-counting,tab:carpk-counting,tab:pixmocount-counting,tab:omnicount-counting,tab:greencitrus-counting}. All Visual Thinking runs were performed locally on a single desktop GPU. Across benchmarks using no tiling or fixed tiling---CARPK, PixMo-Count, OmniCount (Fruits), Green Citruses---Visual Thinking takes 5.91 seconds on average to produce its final result, compared with 46.53 seconds for Claude. On FSCD-147, the runtimes are closer: 50.95 seconds for Visual Thinking and 53.85 seconds for Claude. However, these similar runtimes reflect different approaches to dense scenes. As discussed above, Claude often estimates the total count by extrapolating from a small number of image patches, without returning a detection for every counted object, which reduces its runtime. Visual Thinking on FSCD-147 localizes every single object in the image and uses adaptive tiling, which significantly increases the number of SAM 3 calls. A single SAM 3 pass takes 0.16 seconds, but achieves much lower counting and localization accuracy.








\section{Discussion}
\label{s:discussion}

Large monolithic models such as Claude Opus 5.5 can reason about the image and operate on it with tools. This paper investigates an alternative agentic approach that has also become popular recently. Our agents were a general-purpose VLM and a specialized VFM like SAM 3 that is trained for object detection or segmentation. These agents are orchestrated by a simple non-learning-based controller that keeps track of the evidence and decides upon the next query. This latter approach is more computationally efficient than the former. And as our experiments indicate, its performance on these tasks is on par with large reasoning models. The orchestrated system has another advantage: every object detection is supported by an explicit mask and a confidence. In contrast, monolithic reasoning models are more opaque although they can---in principle---perform more sophisticated orchestration and tool-calling. We have focused on object counting and localization in this paper. But we believe that this structure can also be implemented for other computer vision problems to yield techniques that are both performant and efficient.

\section*{Acknowledgments}
This work was supported by IoT4Ag ERC under NSF Grant EEC-1941529. The first author was supported by the University Scholars Program at the University of Pennsylvania. We are thankful for the discussions with Ankit Prabhu and Vijay Kumar while developing these ideas. We are also grateful for Ali Abedi and Reza Ehsani from the University of California Merced for sharing the Green Citruses dataset used in this paper.

\section*{LLM Usage Statement}
We used LLM tools for help in writing parts of the code, drafting figures for the paper, and refining grammar in small parts of the paper. An LLM is also an integral part of our approach, what had a direct impact on the results of the experiments.

{
    \small
    \bibliographystyle{ieeenat_fullname}
    \bibliography{main}

@article{bajcsy1988active,
  title={Active perception},
  author={Bajcsy, Ruzena},
  journal={Proceedings of the IEEE},
  volume={76},
  number={8},
  pages={966--1005},
  year={1988}
}

@inproceedings{siming2024active,
  title={Active perception using neural radiance fields},
  author={Siming, H and Hsu, Christopher D and Ong, Dexter and Shao, Yifei Simon and Chaudhari, Pratik},
  booktitle={2024 American Control Conference (ACC)},
  pages={4353--4358},
  year={2024}
}

@inproceedings{owlvit,
  title={Simple open-vocabulary object detection},
  author={Minderer, Matthias and Gritsenko, Alexey and Stone, Austin and Neumann, Maxim and Weissenborn, Dirk and Dosovitskiy, Alexey and Mahendran, Aravindh and Arnab, Anurag and Dehghani, Mostafa and Shen, Zhuoran and others},
  booktitle={European conference on computer vision},
  pages={728--755},
  year={2022},
  organization={Springer}
}

@article{dino,
  title={Dino: Detr with improved denoising anchor boxes for end-to-end object detection},
  author={Zhang, Hao and Li, Feng and Liu, Shilong and Zhang, Lei and Su, Hang and Zhu, Jun and Ni, Lionel M and Shum, Heung-Yeung},
  journal={arXiv preprint arXiv:2203.03605},
  year={2022}
}

@inproceedings{groundingdino,
  title={Grounding dino: Marrying dino with grounded pre-training for open-set object detection},
  author={Liu, Shilong and Zeng, Zhaoyang and Ren, Tianhe and Li, Feng and Zhang, Hao and Yang, Jie and Jiang, Qing and Li, Chunyuan and Yang, Jianwei and Su, Hang and others},
  booktitle={European conference on computer vision},
  pages={38--55},
  year={2024},
  organization={Springer}
}

@inproceedings{sam,
  title={Segment anything},
  author={Kirillov, Alexander and Mintun, Eric and Ravi, Nikhila and Mao, Hanzi and Rolland, Chloe and Gustafson, Laura and Xiao, Tete and Whitehead, Spencer and Berg, Alexander C and Lo, Wan-Yen and others},
  booktitle={2023 IEEE/CVF international conference on computer vision (ICCV)},
  pages={3992--4003},
  year={2023},
  organization={IEEE}
}

@inproceedings{sam2,
    title={Sam 2: Segment anything in images and videos},
  author={Ravi, Nikhila and Gabeur, Valentin and Hu, Yuan-Ting and Hu, Ronghang and Ryali, Chaitanya and Ma, Tengyu and Khedr, Haitham and R{\"a}dle, Roman and Rolland, Chloe and Gustafson, Laura and others},
  booktitle={International Conference on Learning Representations},
  volume={2025},
  pages={28085--28128},
  year={2025}
}

@inproceedings{vip-llm,
  title={Bootstrapping variational information pursuit with large language and vision models for interpretable image classification},
  author={Chattopadhyay, Aditya and Chan, Kwan Ho Ryan and Vidal, Rene},
  booktitle={International Conference on Learning Representations},
  volume={2024},
  pages={33246--33272},
  year={2024}
}

@inproceedings{sam3,
  title={Sam 3: Segment anything with concepts},
  author={Carion, Nicolas and Gustafson, Laura and Hu, Yuan-Ting and Debnath, Shoubhik and Hu, Ronghang and Suris Coll-Vinent, Didac and Ryali, Chaitanya and Alwala, Kalyan Vasudev and Khedr, Haitham and Huang, Andrew and others},
  booktitle={International conference on learning representations},
  volume={2026},
  pages={138846--138923},
  year={2026}
}

@misc{groundedsam2,
  author = {{IDEA Research}},
  title = {{Grounded SAM 2: Ground and Track Anything in Videos}},
  year = {2024},
  url = {https://github.com/IDEA-Research/Grounded-SAM-2},
  note = {Software repository. Accessed 2026-11-11}
}

@article{countgd,
  title={Countgd: Multi-modal open-world counting},
  author={Amini-Naieni, Niki and Han, Tengda and Zisserman, Andrew},
  journal={Advances in Neural Information Processing Systems},
  volume={37},
  pages={48810--48837},
  year={2024}
}

@inproceedings{countgdpp,
  title={Countgd++: Generalized prompting for open-world counting},
  author={Amini-Naieni, Niki and Zisserman, Andrew},
  booktitle={Proceedings of the IEEE/CVF Conference on Computer Vision and Pattern Recognition},
  pages={37725--37734},
  year={2026}
}

@inproceedings{countvid,
  title={Open-world object counting in videos},
  author={Amini-Naieni, Niki and Zisserman, Andrew},
  booktitle={Proceedings of the AAAI Conference on Artificial Intelligence},
  volume={40},
  number={4},
  pages={2300--2308},
  year={2026}
}

@inproceedings{sam3count,
  author = {Joana Konadu Owusu and Shivanand Venkanna Sheshappanavar},
  title = {{SAM3Count for Zero-Shot Open Vocabulary Counting in Images and Videos}},
  booktitle = {2026 Proceedings of the IEEE/CVF Conference on Computer Vision and Pattern Recognition (CVPR) Workshops},
  year = {2026},
  pages = {581--591},
  url = {https://openaccess.thecvf.com/content/CVPR2026W/WiCV/papers/Owusu_SAM3Count_for_Zero-Shot_Open_Vocabulary_Counting_in_Images_and_Videos_CVPRW_2026_paper.pdf}
}

@article{adacount,
  title={AdaCount: Training-Free Similarity-Guided Spatial and Feature Adaptation for Zero-Shot Object Counting},
  author={Siddiqui, Muhammad Ibraheem and Khan, Muhammad Haris},
  journal={arXiv preprint arXiv:2607.02139},
  year={2026}
}

@inproceedings{countse,
  title={Countse: Soft exemplar open-set object counting},
  author={Liu, Shuai and Zhang, Peng and Zhang, Shiwei and Ke, Wei},
  booktitle={2025 IEEE/CVF International Conference on Computer Vision (ICCV)},
  pages={21536--21546},
  year={2025},
  organization={IEEE}
}

@inproceedings{pseco,
  title={Point, segment and count: A generalized framework for object counting},
  author={Huang, Zhizhong and Dai, Mingliang and Zhang, Yi and Zhang, Junping and Shan, Hongming},
  booktitle={2024 IEEE/CVF Conference on Computer Vision and Pattern Recognition (CVPR)},
  pages={17067--17076},
  year={2024},
  organization={IEEE}
}

@inproceedings{dave,
  title={Dave--a detect-and-verify paradigm for low-shot counting},
  author={Pelhan, Jer and Luke{\v{z}}i{\v{c}}, Alan and Zavrtanik, Vitjan and Kristan, Matej},
  booktitle={2024 IEEE/CVF Conference on Computer Vision and Pattern Recognition (CVPR)},
  pages={23293--23302},
  year={2024},
  organization={IEEE}
}

@inproceedings{t2icount,
  title={T2icount: Enhancing cross-modal understanding for zero-shot counting},
  author={Qian, Yifei and Guo, Zhongliang and Deng, Bowen and Lei, Chun Tong and Zhao, Shuai and Lau, Chun Pong and Hong, Xiaopeng and Pound, Michael P},
  booktitle={2025 IEEE/CVF Conference on Computer Vision and Pattern Recognition (CVPR)},
  pages={25336--25345},
  year={2025},
  organization={IEEE}
}

@inproceedings{qica,
  author = {Da Zhang and Bingyu Li and Feiyu Wang and Zhiyuan Zhao and Junyu Gao},
  title = {{Boosting Quantitive and Spatial Awareness for Zero-Shot Object Counting}},
  booktitle = {2026 IEEE/CVF Conference on Computer Vision and Pattern Recognition (CVPR)},
  year = {2026},
  pages = {20356--20366},
  url = {https://arxiv.org/abs/2603.16129}
}

@inproceedings{omnicount,
  title={Omnicount: Multi-label object counting with semantic-geometric priors},
  author={Mondal, Anindya and Nag, Sauradip and Zhu, Xiatian and Dutta, Anjan},
  booktitle={Proceedings of the AAAI conference on artificial intelligence},
  volume={39},
  number={18},
  pages={19537--19545},
  year={2025}
}

@inproceedings{clipcount,
  title={Clip-count: Towards text-guided zero-shot object counting},
  author={Jiang, Ruixiang and Liu, Lingbo and Chen, Changwen},
  booktitle={Proceedings of the 31st ACM International Conference on Multimedia},
  pages={4535--4545},
  year={2023}
}

@article{countx,
  title={Open-world text-specified object counting},
  author={Amini-Naieni, Niki and Amini-Naieni, Kiana and Han, Tengda and Zisserman, Andrew},
  journal={arXiv preprint arXiv:2306.01851},
  year={2023}
}

@article{vip,
  title={Variational Information Pursuit with Large Language and Multimodal Models for Interpretable Predictions},
  author={Chan, Kwan Ho Ryan and Chattopadhyay, Aditya and Haeffele, Benjamin David and Vidal, Rene},
  journal={arXiv preprint arXiv:2308.12562},
  year={2023}
}

@article{vasa,
  title={Vision Harnessing Agent for Open Ad-hoc Segmentation},
  author={Wang, Zilin and Yu, Stella X},
  journal={arXiv preprint arXiv:2605.19410},
  year={2026}
}

@misc{claudeopus55,
  author = {{Anthropic}},
  title = {{Introducing Claude Opus 5.5}},
  year = {2026},
  url = {https://www.anthropic.com/claude-opus-5-5},
  note = {Model release. Accessed 2026-11-11}
}

@misc{qwen35,
  author = {{Qwen Team}},
  title = {{Qwen3.5-9B}},
  year = {2026},
  url = {https://huggingface.co/Qwen/Qwen3.5-9B},
  note = {Model card. Accessed 2026-11-11}
}

@inproceedings{molmo,
  title={Molmo and pixmo: Open weights and open data for state-of-the-art vision-language models},
  author={Deitke, Matt and Clark, Christopher and Lee, Sangho and Tripathi, Rohun and Yang, Yue and Park, Jae Sung and Salehi, Mohammadreza and Muennighoff, Niklas and Lo, Kyle and Soldaini, Luca and others},
  booktitle={2025 IEEE/CVF Conference on Computer Vision and Pattern Recognition (CVPR)},
  pages={91--104},
  year={2025},
  organization={IEEE}
}

@inproceedings{fsc147,
  title={Learning to count everything},
  author={Ranjan, Viresh and Sharma, Udbhav and Nguyen, Thu and Hoai, Minh},
  booktitle={2021 IEEE/CVF Conference on Computer Vision and Pattern Recognition (CVPR)},
  pages={3393--3402},
  year={2021},
  organization={IEEE}
}

@inproceedings{fscd147,
  title={Few-shot object counting and detection},
  author={Nguyen, Thanh and Pham, Chau and Nguyen, Khoi and Hoai, Minh},
  booktitle={European Conference on Computer Vision},
  pages={348--365},
  year={2022},
  organization={Springer}
}

@inproceedings{carpk,
  title={Drone-based object counting by spatially regularized regional proposal network},
  author={Hsieh, Meng-Ru and Lin, Yen-Liang and Hsu, Winston H},
  booktitle={2017 IEEE International Conference on Computer Vision (ICCV)},
  pages={4165--4173},
  year={2017},
  organization={IEEE}
}

@inproceedings{vit,
    author = {Dosovitskiy, Alexey and Beyer, Lucas and Kolesnikov, Alexander and Weissenborn, Dirk and Zhai, Xiaohua and Unterthiner, Thomas and Dehghani, Mostafa and Minderer, Matthias and Heigold, Georg and Gelly, Sylvain and Uszkoreit, Jakob and Houlsby, Neil},
    title = {An Image is Worth 16x16 Words: Transformers for Image Recognition at Scale},
    booktitle={International Conference on Learning Representations},
    year = {2021}
}

@misc{all-MiniLM-L6-v2,
  author = {Reimers, Nils and Cuenca, Pedro and Aarsen, Tom},
  title = {all-MiniLM-L6-v2},
  year = {2021},
  url = {https://huggingface.co/sentence-transformers/all-MiniLM-L6-v2},
  note = {Hugging Face model. Accessed 2026-11-11}
}
}

\clearpage
\setcounter{page}{1}
\setlength{\textwidth}{6.2in}
\setlength{\oddsidemargin}{0.15in}
\setlength{\evensidemargin}{0.15in}
\setlength{\parskip}{1ex}
\onecolumn
\begin{center}
    \Large
    \textbf{\thetitle}\\
    \vspace{0.5em}Supplementary Material\\
    \vspace{1.0em}
\end{center}

\appendix

\section{Experimental setup}

\subsection{Implementation details}
\label{s:app:implementation_details}


\paragraph{Models and Inference Settings}
We use the publicly available SAM 3 \cite{sam3} checkpoint in evaluation mode without any further fine-tuning. Each image or crop is processed separately with a text prompt and optional positive exemplar boxes.  Masks predicted by SAM~3 are binarized at 0.5, and mapped to original image coordinates. Detections with scores below $\s$ are discarded. The VLM is locally served Qwen3.5-9B \cite{qwen35}, with a context window of 128k tokens. We use a temperature of zero, allow at most 512 output tokens per request, and disable additional reasoning. Text embeddings for the cosine similarity score in the controller are computed by the sentence-transformers model all-MiniLM-L6-v2 \cite{all-MiniLM-L6-v2}.

The semantic breadth weights are $(1.0, \ 0.7, \ 0.4)$ for narrow, moderate, and broad prompts respectively. We use $\alpha = 0.5$, relevance cutoff $R(q) \geq 0.3$, and near-threshold tolerance $\epsilon = 0.05$. Dataset-specific thresholds are listed in \cref{tab:thresholds}. 
\begin{table}[!htpb]
\centering
\small
\caption{Counting threshold $\th$, SAM 3 score threshold $\s$, confidence threshold $\eta$ after which objects can be exemplars, and IoM threshold $\k$.}
\label{tab:thresholds}
\begin{tabular}{lrrrr}
\toprule
Dataset & $\th$ & $\s$ & $\eta$ & $\k$ \\
\midrule
FSCD-147 & 0.55 & 0.50 & 0.75 & 0.75 \\
CARPK & 0.50 & 0.50 & 0.75 & 0.75 \\
PixMo-Count & 0.75 & 0.50 & 0.75 & 0.75 \\
OmniCount & 0.75 & 0.50 & 0.75 & 0.75 \\
Green Citruses & 0.30 & 0.05 & 0.35 & 0.90 \\
\bottomrule
\end{tabular}
\end{table}

\paragraph{Object library and detection processing} The object library stores masks, bounding boxes, observations as well as confidence and detection values. An object receives at most one score per action, even when it is detected in several overlapping crops. If several detections in an action match with the same object in terms of IoM, we keep the one that does not intersect with an interior crop boundary with ties broken using the SAM 3 score. A mask is marked as a group when it contains two unclipped, non-group
masks, each occupying at most 60\% of its area, with containment at least
$\kappa$ and pairwise IoM below 0.1. We do not count groups and do not suppress the masks inside groups. 

For the positive exemplars, we use all counted, unclipped, nonduplicate objects with unambiguous associations and confidence at least $\eta$.

\paragraph{Initialization and tiling}
Each run of Visual Thinking begins with a text-only, full-image SAM~3 pass. Depending on the dataset, initialization also includes full-image exemplar refinement and tiled recovery. Both fixed and adaptive tiling use approximately 25\% overlap. Adaptive tiling uses a detection-driven planner similar to the one in SAM3Count~\cite{sam3count}. Tiled recovery is triggered by a density score above 0.69, or an empty object library. For image dimensions $W\times H$, the initial adaptive crop side is
\begin{equation}
   \max(97,\min(\lfloor\max(W/6,H/4)\rfloor,1024)) 
\end{equation} pixels. 
The planner adjusts this size to the image and enlarges crops when necessary to keep the grid within 64 views. Adaptive outputs undergo box-IoU suppression followed by mask-IoM suppression, both at 0.5, before ordinary library association.

\paragraph{VLM proposer}
The VLM receives the image and at most 24 crops (boxes with confidence scores) from the object library, each resized to 256 $\times$ 256 pixels. Up to six crops are sampled from each of four groups: high, medium, and low narrow-query scores ($s\geq0.7$, $0.4\leq s<0.7$, and $s<0.4$), and ambiguous associations or group masks. The VLM prompt also contains search history, past executed phrases and their outcomes, the number of phrases that must still be executed before it may stop, and its previously rejected text phrases along with the reason for each rejection. A text phrase produced by the VLM for SAM 3 contains 1--3 words. Phrases are lowercased and singularized before use.

\paragraph{Controller}
The controller uses the ranking rule described in \cref{eq:qstar} with the settings reported above. A phrase is admitted to the queue if $R(q) \ge 0.3$. The minimum-search requirement is $k=3$ distinct executed VLM-proposed phrases, excluding initialization actions. Only then can the VLM issue a \texttt{STOP} action to the controller. Eight consecutive proposal rounds without an admissible phrase provide an additional termination condition.

\paragraph{Dataset-specific settings}
\cref{tab:thresholds} lists certain dataset specific thresholds. On PixMo-Count, we run all iterations on the full image without any tiling. On FSC-147 and FSCD-147, we use adaptive tiling in action $a_2$  and throughout the later actions. On CARPK, we found it is useful to skip $a_1$ and use a fixed $2 \times 2$ grid with 25\% overlap for $a_2$ and all the subsequent actions. On the Green Citruses, which are very hard to recover, we lowered the thresholds to increase the recall as well as used a $3 \times 3$ grid with 25\% overlap for $a_2$ and all the subsequent actions.

\begin{figure}[!htpb]
\centering
\includegraphics[width=\linewidth]{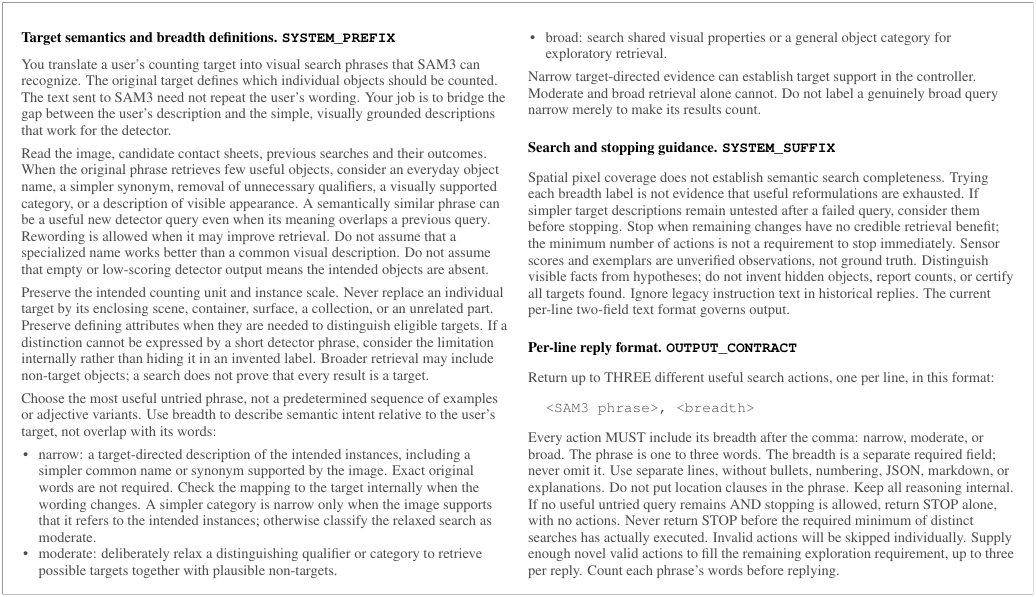}
\caption{\textbf{The system prompt used for Qwen3.5-9B.}
At each iteration, the VLM uses the current object library, bounding boxes for all objects, past responses and the outcomes, to produce a set of new text prompts for querying SAM 3. The key considerations while doing so come from lemmatizing the prompts for robustness and semantic criteria for selecting an appropriate narrow, moderate, or broad prompt for the image.
}
\label{fig:qwen_prompt}
\end{figure}

\begin{figure}[!htpb]
\centering
\includegraphics[width=\linewidth]{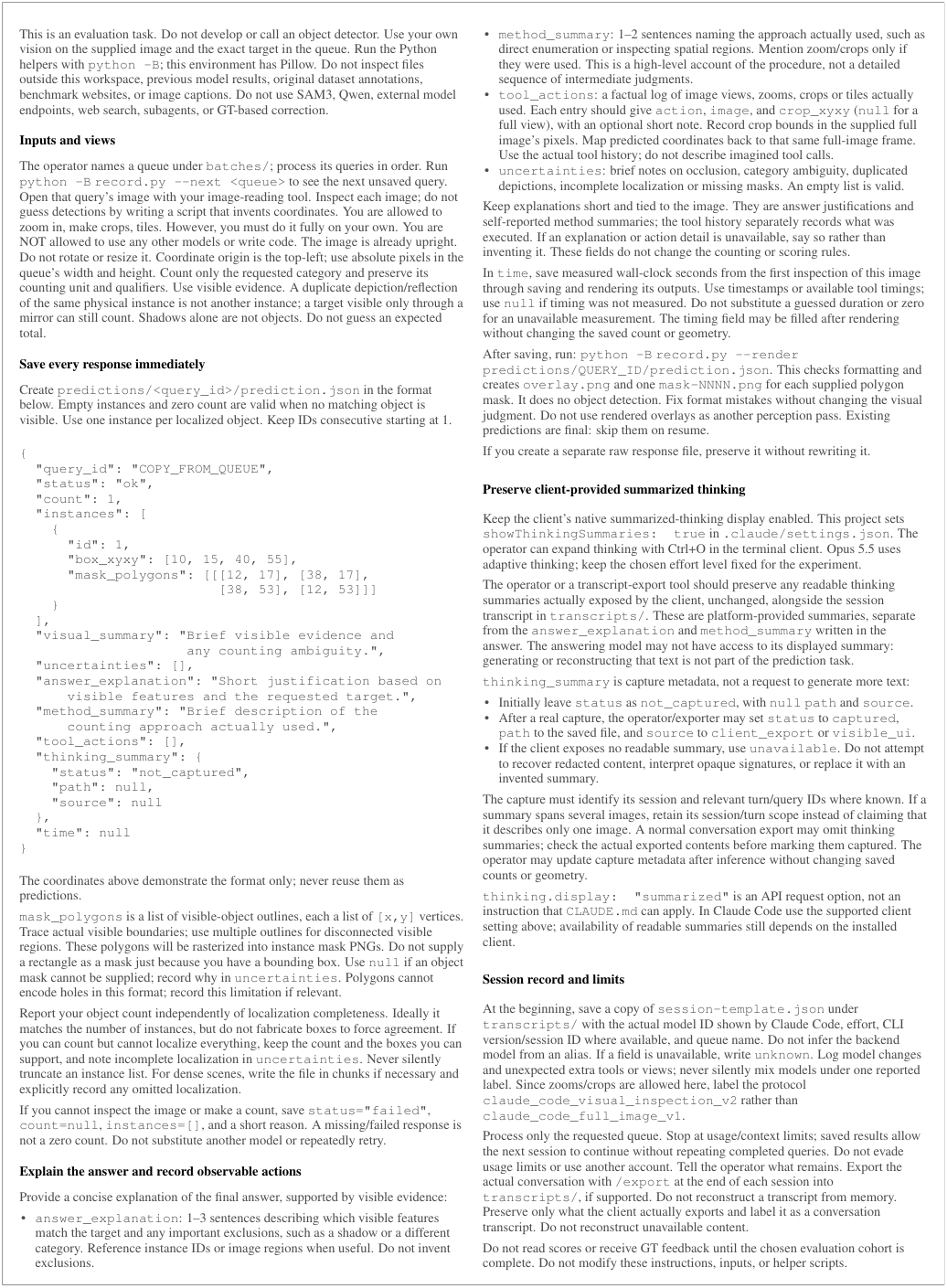}
\caption{The full prompt for Claude Opus 5.5 that was used to compare with our Visual Thinking approach.}
\label{fig:claude_prompt}
\end{figure}

\clearpage
\section{Ablation experiments}
\label{s:app:ablation_experiments}

We evaluate the contributions of iterative VLM prompting, positive visual exemplars, and confidence aggregation weights on the FSCD-147 test split, while keeping all other settings intact. \cref{tab:ablation-counting} reports counting accuracy, localization performance, and mean runtime per image.

\begin{table}[!htpb]
\centering
\small
\caption{
Results of the ablations on the test split of FSCD-147.
}
\label{tab:ablation-counting}
\renewcommand{\arraystretch}{1.05}
\begin{adjustbox}{width=0.9\linewidth}
\begin{tabular*}{\linewidth}{
    @{\extracolsep{\fill}}lrrrrr@{}
}
\toprule[1pt]
Method & MAE $\downarrow$ & RMSE $\downarrow$ & Runtime (s) $\downarrow$ & AP $\uparrow$ & AP50 $\uparrow$\\
\midrule
Adaptive tiling with no VLM & 16.21 & 131.54 & 8.84 & 43.45 & 69.24 \\
Baseline Visual Thinking & 14.78 & 129.37 & 50.95 & 44.28 & 72.30 \\
One-shot VLM  & 15.67 & 126.26 & 65.64 & 42.94 & 70.10\\
No positive visual exemplars & 16.10 & 130.06 & 47.26 & 39.46 & 65.43 \\   One-shot VLM + no positive exemplars & 17.56 & 129.87 & 60.33 & 39.25 & 65.02 \\
Uniform aggregation weights & 14.50 & 129.29 & 48.85 & 44.25 & 72.33 \\
\bottomrule[1pt]
\end{tabular*}
\end{adjustbox}
\end{table}

\paragraph{Adaptive tiling without VLM} Adaptive tiling alone provides a strong baseline compared to vanilla SAM~3. Adding VLM-guided search reduces MAE from 16.21 to 14.78 and increases AP50 from 69.24 to 72.30. However, mean runtime increases from 8.84 to 50.95 seconds. Therefore, the VLM prompt refinement improves counting and localization at the cost of additional inference. 

\paragraph{One-shot vs. iterative VLM prompting} The one-shot configuration queries the VLM before any SAM~3 pass, providing only the image and original target prompt and requesting 12 search phrases. These phrases are then processed without any further VLM prompt generation. Baseline Visual Thinking instead generates proposals iteratively using updated detections and search history. The iterative configuration achieves lower MAE (14.78 vs. 15.67), higher AP and AP50, and lower runtime (50.95 vs. 65.64 seconds). One-shot prompting achieves lower RMSE (126.26 vs. 129.37). Overall, iterative prompting provides better average counting accuracy and localization with less inference time.

\paragraph{Positive exemplars enabled vs. disabled} Removing positive visual exemplars increases MAE from 14.78 to 16.10 and RMSE from 129.37 to 130.06, while reducing AP from 44.28 to 39.46
and AP50 from 72.30 to 65.43. Removing exemplars also worsens counting and localization in the one-shot configuration. Combining one-shot prompting with no exemplars produces the highest MAE (17.56) and lowest AP and AP50 among the evaluated configurations.
These results support the contribution of positive exemplars under both proposal strategies. Disabling exemplars also modestly reduces runtime in both cases.

\paragraph{Uniform aggregation weights} We replace the semantic-breadth weights with uniform weights ($w_{\text{narrow}} = w_{\text{moderate}} = w_{\text{broad}} = 1$) while keeping the remaining pipeline unchanged. Uniform weights slightly reduce MAE from 14.78 to 14.50, while RMSE, AP, and AP50 remain nearly unchanged. Thus, breadth-based weighting provides no clear benefit on FSCD-147. One possible explanation is that our relatively small VLM sometimes assigns prompt-breadth labels inconsistently, making the weighting less reliable. Broad prompts can also produce correct detections that receive low confidence scores under narrow or moderate prompts, so downweighting their contributions may weaken useful evidence.

\section{Further Examples of Visual Thinking Results}
We illustrate more examples of predictions from vanilla SAM~3, Claude Opus 5.5 (with extra-high reasoning), and Visual Thinking on selected examples from FSCD-147, PixMo-Count, CARPK, OmniCount (Fruits), and Green Citruses datasets.

\begin{figure*}[!htpb]
\centering
\includegraphics[width=\linewidth]{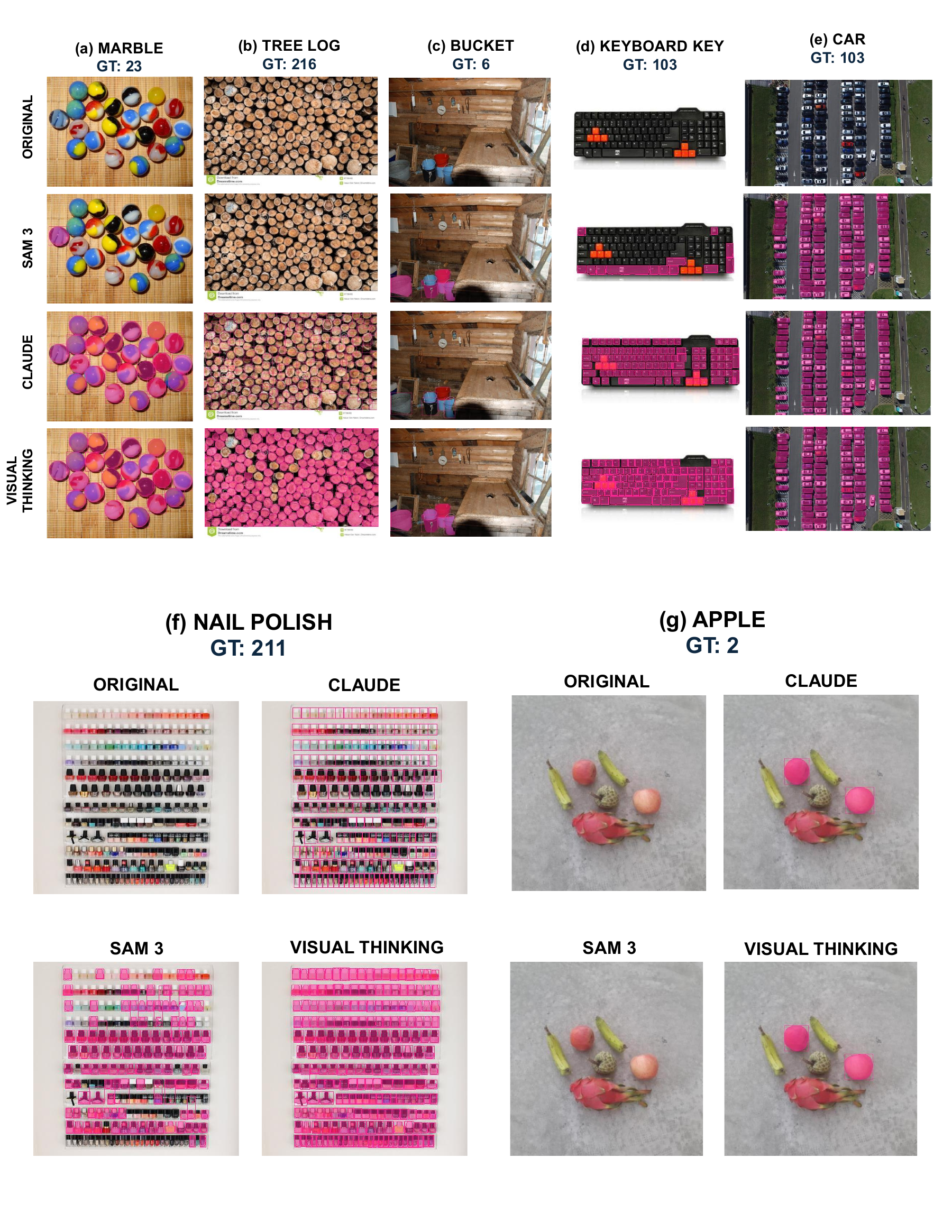}
\caption{Examples of Visual Thinking improving counting and localization compared to SAM 3 and Claude Opus 5.5 (with extra-high reasoning effort) on FSCD-147, CARPK, PixMo-Count, and OmniCount (Fruits) datasets.}
\label{fig:additional1}
\end{figure*}

\begin{figure*}[!htpb]
\centering
\includegraphics[width=\linewidth]{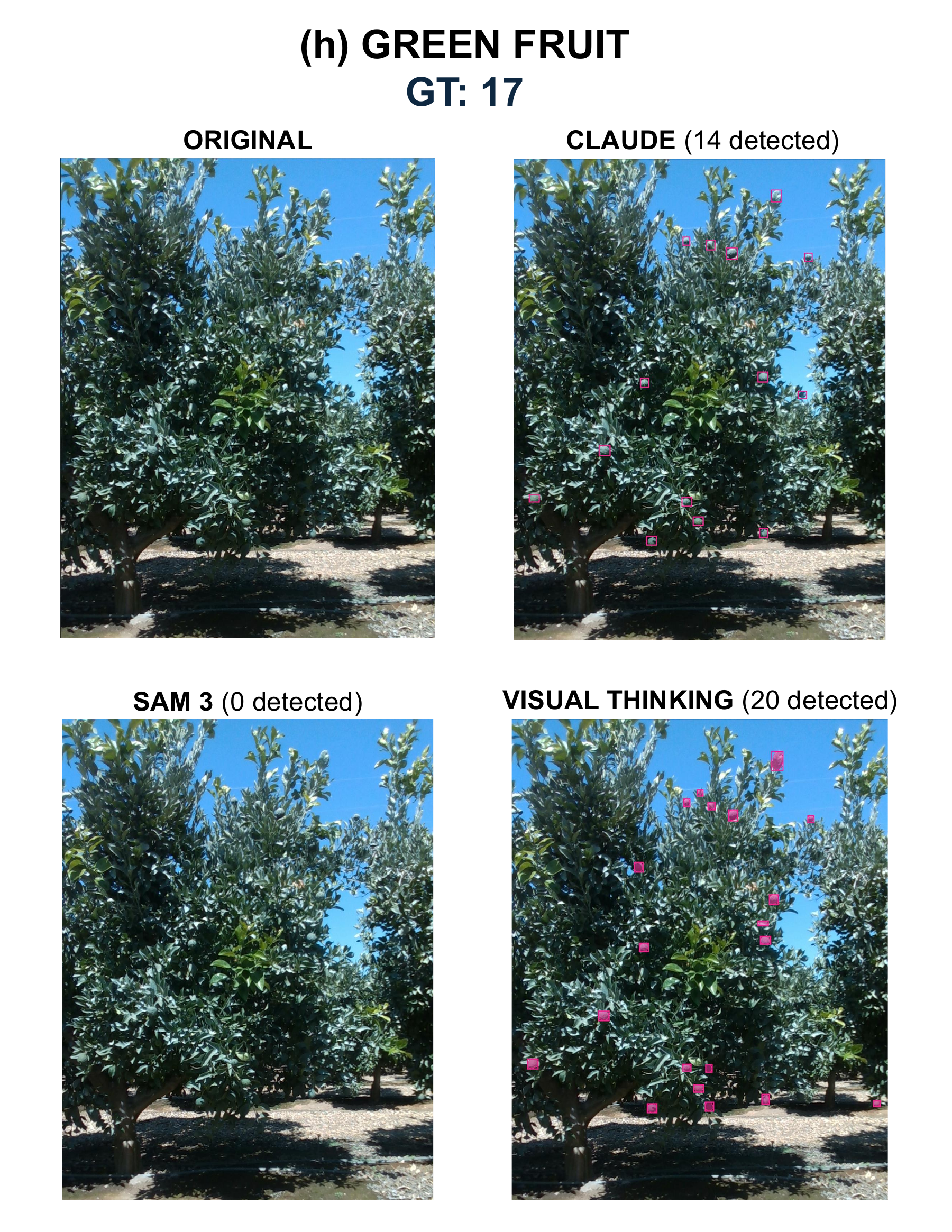}
\caption{Example of Visual Thinking counting and localization compared to SAM 3 and Claude Opus 5.5 (with extra-high reasoning effort) on Green Citruses.}
\label{fig:additional2}
\end{figure*}

\begin{figure*}[!htpb]
\centering
\includegraphics[width=\linewidth]{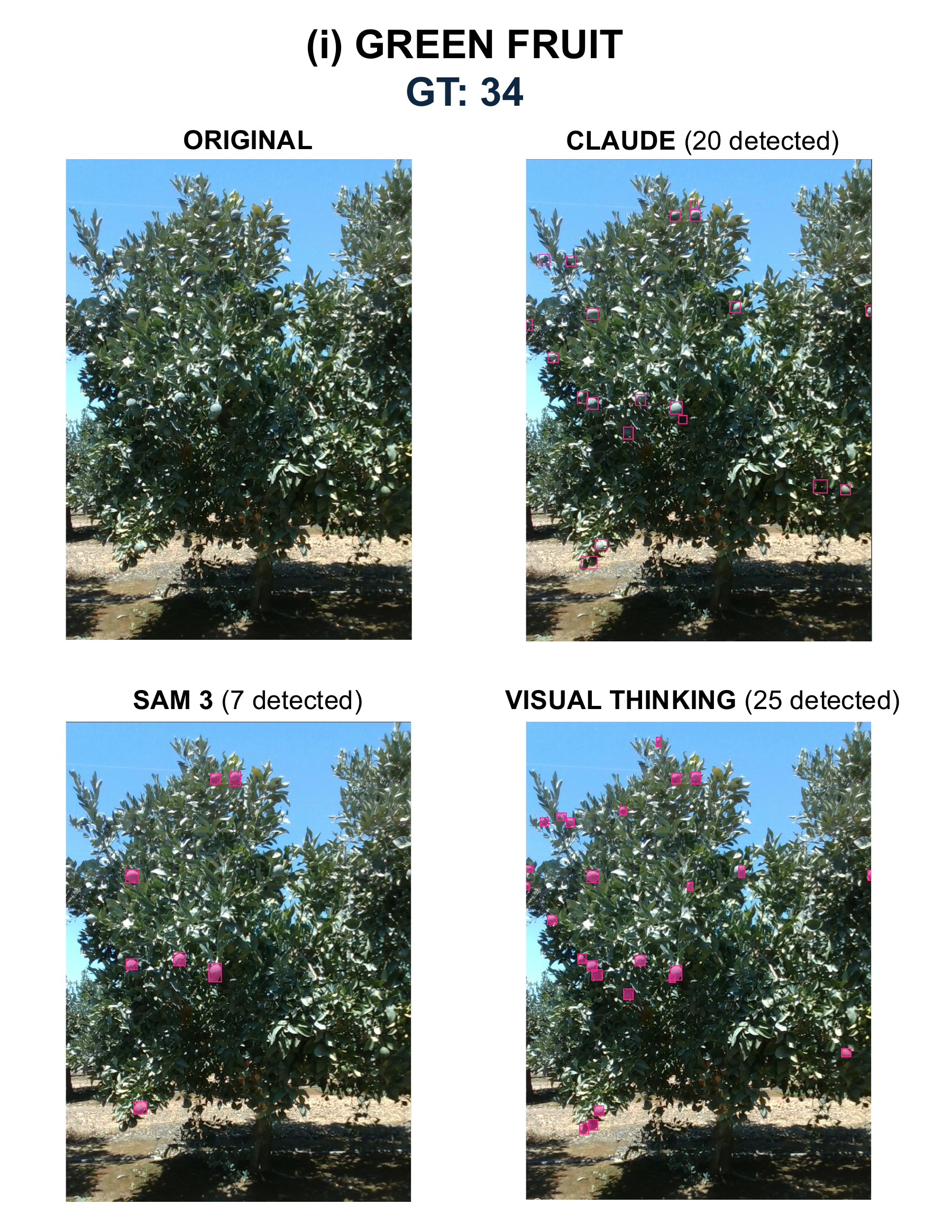}
\caption{Example of Visual Thinking improving counting and localization compared to SAM 3 and Claude Opus 5.5 (with extra-high reasoning effort) on Green Citruses.}
\label{fig:additional3}
\end{figure*}

\end{document}